\documentclass{article}

\usepackage{PRIMEarxiv}

\usepackage[utf8]{inputenc} 
\usepackage[T1]{fontenc}    
\usepackage{hyperref}       
\usepackage{url}            
\usepackage{booktabs}       
\usepackage{amsfonts}       
\usepackage{nicefrac}       
\usepackage{microtype}      
\usepackage{lipsum}
\usepackage{fancyhdr}       
\usepackage{graphicx}       
\graphicspath{{media/}}     
\usepackage{amsmath}
\usepackage{caption}     
\usepackage{subcaption}  
\usepackage{tabularx}    
\usepackage{colortbl}
\usepackage{multirow}
\usepackage{bm}
\usepackage[most]{tcolorbox}

\newcolumntype{Y}{>{\centering\arraybackslash}X}

\usepackage[colorinlistoftodos]{todonotes}

\usepackage{tikz}
\usetikzlibrary{shapes.geometric, arrows}
\usetikzlibrary{arrows} 

\tikzstyle{startstop} = [cylinder, shape border rotate=90, aspect=.1, draw, minimum height=.6cm, minimum width=.6cm, text width=1.35cm, align=center]
\tikzstyle{process} = [rectangle, draw, fill=gray!20, minimum width=1.2cm, minimum height=.6cm, text width=1.2cm, align=center]
\tikzstyle{decision} = [diamond, draw, minimum width=.6cm, minimum height=.6cm]
\tikzstyle{arrow} = [thick, ->, >=stealth]
\tikzstyle{line} = [thick, -, >=stealth]
\tikzstyle{box} = [rectangle, draw, minimum width=1.2cm, minimum height=.4cm, text width=1.5cm, align=center]

\usetikzlibrary{calc}
\usetikzlibrary{arrows.meta, positioning}
\usepackage{float}
\title{Neurosymbolic AI Transfer Learning Improves Network Intrusion Detection}

\author{
Huynh T. T. Tran, Jacob Sander, Achraf Cohen, Brian Jalaian \\
\textit{University of West Florida, Pensacola, USA} \\
\texttt{\{ht68, jhs39\}@students.uwf.edu}, \texttt{\{acohen, bjalaian\}@uwf.edu} \\
\and
\textbf{Nathaniel D. Bastian} \\
\textit{United States Military Academy, West Point, New York, USA} \\
\texttt{nathaniel.bastian@westpoint.edu}
}

\begin{document}

\maketitle

\begin{abstract}
Transfer learning is commonly utilized in various fields such as computer vision, natural language processing, and medical imaging due to its impressive capability to address subtasks and work with different datasets. However, its application in cybersecurity has not been thoroughly explored. In this paper, we present an innovative neurosymbolic AI framework designed for network intrusion detection systems, which play a crucial role in combating malicious activities in cybersecurity. Our framework leverages transfer learning and uncertainty quantification. The findings indicate that transfer learning models, trained on large and well-structured datasets, outperform neural-based models that rely on smaller datasets, paving the way for a new era in cybersecurity solutions.
\end{abstract}


\section{Introduction}
Network Intrusion Detection Systems (NIDS) play a critical role in modern cybersecurity by monitoring network traffic to detect and classify malicious activities \cite{al2021intelligent}. Traditional machine learning models, while effective in many scenarios, struggle to address the increasingly complex and diverse nature of cyber threats. To overcome these challenges, Neurosymbolic AI (NSAI) has emerged as a promising approach. NSAI combines the powerful data-processing capabilities of neural networks with the logical reasoning strengths of symbolic AI \cite{sander2024uncertainty,jalaian2023neurosymbolic}. The symbolic component, often implemented through models such as XGBoost, leverages rule-based decision-making, thereby enhancing the overall robustness of intrusion detection systems.

However, the limitation of current NSAI systems is their reliance on the availability of labeled data tailored to specific datasets or attack types. As new attacks and datasets emerge, building a new model for each scenario can be resource-intensive and inefficient. This challenge is particularly evident in NIDS, where different datasets and tasks may require tailored models, hindering scalability and adaptability.

Transfer learning can be considered a viable solution. Transfer learning leverages pre-trained models, adapting them to new tasks or domains with limited labeled data. While widely used in fields such as computer vision \cite{shahoveisi2023application}, natural language processing \cite{amiriparian2022deepspectrumlite, moon2014multimodal}, and medical imaging \cite{kim2022transfer}, transfer learning remains underexplored in the cybersecurity domain. 

This work extends our Open Set Recognition with Deep Embedded Clustering for XGBoost and Uncertainty Quantification (ODXU) framework \cite{sander2024uncertainty}, by integrating transfer learning to enhance scalability and adaptability. \textcolor{black}{Using the Canadian Institute for Cybersecurity Intrusion Detection Systems 2017 (CIC-IDS-2017) dataset \cite{sharafaldincicids2017}, which is large and resembles real-world scenarios. This dataset contains approximately 7 million samples of both benign traffic and up-to-date common cyberattacks. Our contributions are:}

\begin{itemize}
    \item Developing a transfer learning framework to adapt pre-trained NSAI models to new cybersecurity datasets from the Army Cyber Institute, enabling improved adaptability and generalization in dynamic threat environments.
    \item Integrating and evaluating more uncertainty quantification techniques, such as SHAP value and information gain, to enhance the interpretability and reliability of model predictions in uncertain conditions.
\end{itemize}

The paper is organized as follows: Section \ref{sec:background} reviews related work, Section \ref{sec:methodology} outlines the ODXU model and transfer learning framework, Section \ref{sec:exp_setup} details the experimental setup, Section \ref{sec:results} presents results and discussion, and Section \ref{sec:conclusion} concludes with future work.

\begin{figure}[t!]
\centering
\tiny
\begin{tikzpicture}[node distance=0cm]

    \node (start) [startstop] {Raw Payload};
    \node (byte) [process, right of=start, xshift=1.62cm] {Payload-Byte};
    \node (payload) [startstop, right of=byte, xshift=1.62cm] {1500-byte Payload};
    \node (dec) [process, right of=payload, xshift=1.62cm] {DEC};
    \node (latent) [startstop, right of=dec, xshift=1.62cm] {\hspace{.05cm} Latent \hspace{.3cm} representations};

    \node (inter) [coordinate] at ($(latent.east)+(.1cm,0)$) {};

    \node (line1) [coordinate] at ($(inter.north)+(0,.6cm)$) {};
    \node (uq) [process, right of=line1, xshift=.9cm] {UQ \hspace{2cm} Metamodel};
    
    \node (inter1) [coordinate] at ($(uq.east)+(.2cm,0)$) {};
    \node (line2) [coordinate] at ($(inter1.north)+(0,.4cm)$) {};
    \node (line3) [right of=line2, xshift=1.5cm] {};
    \node (text1) [above of=line3, xshift = -.8cm, yshift = .15cm] {\textit{high certainty score}};
    \node (high) [right of=line3, xshift=1.15cm, text width=2cm] {Trust the XGBoost classification results};

    \node (line4) [coordinate] at ($(inter1.north)+(0,-.4cm)$) {};
    \node (line5) [right of=line4, xshift=1.5cm] {};
    \node (text2) [below of=line5, xshift = -.9cm, yshift = -.15cm] {\textit{low certainty score}};
    \node (low) [right of=line5, xshift=1.15cm, text width=2.4cm] {\hspace{.2cm} Suspicious packet \hspace{.2cm} (misclassified or unknown)};

    \node (line6) [coordinate] at ($(inter.north)+(0,-.6cm)$) {};
    \node (line7) [right of=line6, xshift=.6cm] {};
    \node (xgb) [process, right of=line7, xshift=.3cm] {XGBoost};

    \node (inter2) [coordinate] at ($(xgb.east)+(.2cm,0)$) {};
    \node (line8) [coordinate] at ($(inter2.north)+(0,.3cm)$) {};
    \node (line9) [right of=line8, xshift=1.5cm] {};  
    \node (benign) [right of=line9,  xshift=1.1cm] {Benign};

    \node (line10) [coordinate] at ($(inter2.south)+(0,-.3cm)$) {};
    \node (line11) [right of=line10, xshift=1.5cm] {};  
    \node (attack) [right of=line11,  xshift=1.1cm] {One of the known attacks};

    \draw [arrow] (start) -- (byte);
    \draw [arrow] (byte) -- (payload);
    \draw [arrow] (payload) -- (dec);
    \draw [arrow] (dec) -- (latent);

    \draw [line]  (latent) -- (inter);
    
    \draw [line] (inter) -- (line1);
    \draw [arrow] (line1) -- (uq);

    \draw [line] (uq) -- (inter1);
    \draw [line] (inter1) -- (line2);
    \draw [line] (line2) -- (line3);

    \draw [line] (inter1) -- (line4);
    \draw [line] (line4) -- (line5);
    
    \draw [line] (inter) -- (line6);
    \draw [arrow] (line6) -- (xgb);

    \draw [line] (xgb) -- (inter2);
    \draw [line] (inter2) -- (line8);
    \draw [line] (line8) -- (line9);

    \draw [line] (inter2) -- (line10);
    \draw [line] (line10) -- (line11);

\end{tikzpicture}
\vspace{-0.5em}
\caption{The architecture of the ODXU model \cite{sander2024uncertainty}}
\label{fig:pipeline}
\end{figure}
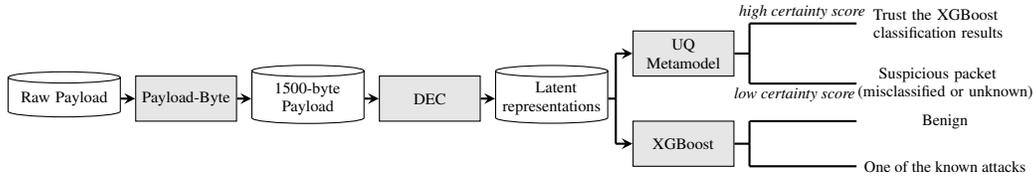

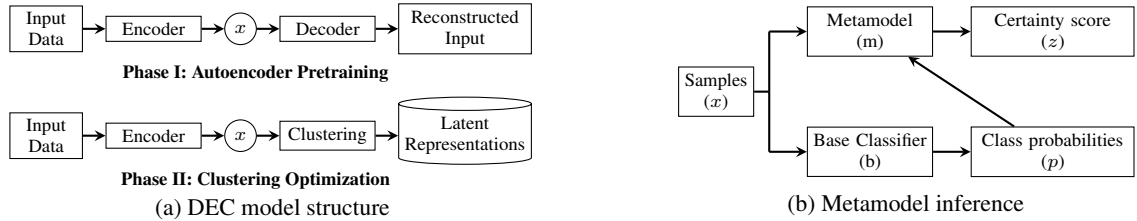
\begin{figure}[t!]
    \centering
    \begin{subfigure}[b]{.52\textwidth}
    \begin{minipage}{\textwidth}
        \centering
        \scriptsize
        \begin{tikzpicture}[node distance=0cm, every node/.style={align=center}]

        \node (input1) [draw, rectangle, text width = .8cm] {Input \\ Data};
        \node (encoder1) [draw, rectangle, right=.3cm of input1, text width = 1.1cm] {Encoder};
        \node (latent1) [draw, circle, right=.3cm of encoder1] {$x$};
        \node (decoder) [draw, rectangle, right=.3 cm of latent1, text width = 1.1cm] {Decoder};
        \node (recon) [draw, rectangle, right=.3 cm of decoder, , text width = 1.6cm] {Reconstructed \\ Input};

        \draw[arrow] (input1) -- (encoder1);
        \draw[arrow] (encoder1) -- (latent1);
        \draw[arrow] (latent1) -- (decoder);
        \draw[arrow] (decoder) -- (recon);

        \node at ($(input1)!.5!(recon)+(0,-.6cm)$) {\textbf{Phase I: Autoencoder Pretraining}};

        \node (input2) [draw, rectangle, below=.8cm of input1, text width = .8cm] {Input \\ Data};
        \node (encoder2) [draw, rectangle, right=.3cm of input2, text width = 1.1cm] {Encoder};
        \node (latent2) [draw, circle, right=.3cm of encoder2] {$x$};
        \node (cluster) [draw, rectangle, right=.3cm of latent2, text width = 1.1cm] {Clustering};
        \node (latent) [startstop, right=.3cm of cluster, , text width = 1.6cm] {Latent \\ Representations};

        \draw[arrow] (input2) -- (encoder2);
        \draw[arrow] (encoder2) -- (latent2);
        \draw[arrow] (latent2) -- (cluster);
        \draw[arrow] (cluster) -- (latent);

        \node at ($(input2)!.5!(latent)+(0,-.6cm)$) {\textbf{Phase II: Clustering Optimization}};

    \end{tikzpicture}
    \vspace{-.2cm}
        \caption{DEC model structure}
        \label{fig:dec_model}
    \end{minipage}
    \end{subfigure}
    \hfill
    \begin{subfigure}[b]{.46\linewidth}
    \begin{minipage}{\textwidth}
        \centering
        \scriptsize
        \begin{tikzpicture}[node distance=0, every node/.style={align=center}]

            \node (samples) [draw, rectangle] {Samples \\ ($x$)};
            \node (inter) [coordinate] at ($(samples.east)+(.2cm,0)$) {};
            \node (line1) [coordinate] at ($(inter.north)+(0cm,.8cm)$) {};
            \node (metamodel) [draw, rectangle, right=.5cm of line1, text width = 1.5cm] {Metamodel \\ (m)};
            \node (zscore) [draw, rectangle, right=.5cm of metamodel, text width = 2cm] {Certainty score \\ ($z$)};

            \node (line2) [coordinate] at ($(inter.south)+(0,-.8cm)$) {};
            \node (basemodel) [draw, rectangle, right=.5cm of line2, text width = 1.5cm] {Base Classifier \\ (b)};
            \node (probability) [draw, rectangle, right=.5cm of basemodel, text width = 2cm] {Class probabilities \\ ($p$)};

            \draw [line] (samples) -- (inter);
            \draw [line] (inter) -- (line1);
            \draw [arrow] (line1) -- (metamodel);
            \draw [arrow] (metamodel) -- (zscore);

            \draw [line] (inter) -- (line2);
            \draw [arrow] (line2) -- (basemodel);
            \draw [arrow] (basemodel) -- (probability);

            \draw [arrow] (probability) -- (metamodel);
            
        \end{tikzpicture}
        \caption{Metamodel inference}
        \label{fig:metamodel}
    \end{minipage}
    \end{subfigure}
    \caption{General structure of (a) DEC and (b) metamodel}
    \label{fig:general_structure}
\end{figure}

\section{Background and Related Work}
\label{sec:background}

\subsection{Network Intrusion Detection Systems}

Network Intrusion Detection Systems (NIDS) are central to modern cybersecurity frameworks, monitoring network traffic to detect and mitigate malicious activities. Traditional NIDS primarily rely on signature-based methods, which compare observed traffic against predefined patterns of known attacks \cite{al2021intelligent, sander2024uncertainty}. While effective for known threats, these systems often fail against novel or unknown attacks due to their static nature and reliance on comprehensive signature databases.

To address these limitations, data-driven anomaly detection has emerged as a more adaptive alternative. These methods utilize machine learning models, including neural networks, decision trees, and ensemble techniques such as XGBoost, to learn normal traffic patterns and identify deviations that may indicate intrusions \cite{al2021intelligent}. Such models improve generalization and are better suited for detecting emerging threats in dynamic network environments.

\subsection{Transfer Learning with Neurosymbolic AI}

Transfer learning is widely adopted across domains to improve model performance when labeled data for the target domain is limited or distribution shifts occur. In computer vision tasks, surgical fine-tuning—selective adjustment of specific neural network layers—has shown significant improvement in adapting models to input-level, feature-level, and output-level shifts \cite{lee2022surgical, shahoveisi2023application}. Techniques such as TransTailor have optimized model architectures through targeted pruning to better align the model structure with task-specific data distributions \cite{liu2021transtailor, moon2014multimodal}. Adaptive fine-tuning approaches, like SpotTune, dynamically decide which layers to fine-tune per input instance, enhancing flexibility and performance \cite{guo2019spottune, moon2014multimodal}.

Although effective in various fields, transfer learning techniques are still underutilized in cybersecurity applications. Meanwhile, NSAI has emerged as a promising approach that combines the predictive capabilities of neural networks with the interpretability of symbolic reasoning \cite{sander2024uncertainty}.

In this paper, we propose integrating transfer learning within the NSAI framework. By allowing the reuse and adaptation of pre-trained NSAI models for new datasets or related tasks, this approach improves the scalability and robustness of intrusion detection systems in dynamic threat environments.

\subsection{Uncertainty Quantification}

Uncertainty Quantification (UQ) is crucial in security-critical systems, where reliable decision-making relies on confidence in model predictions. UQ techniques estimate prediction uncertainty, helping identify unknowns or out-of-distribution inputs. Standard methods for assessing the reliability of predictions from classifiers include confidence scores, Shannon Entropy \cite{smith2011quantifying}, and BlackBox metamodeling \cite{chen2019ibmwhitebox}. The latter involves training a secondary model to evaluate the trustworthiness of the outputs from the primary classifier \cite{sander2024uncertainty}. These techniques improve the robustness of NIDS by identifying uncertain predictions that may indicate new types of threats.

Recent advancements, such as SHAP \cite{lundberg2017unified} (Shapley Additive Explanations) and Information Gain \cite{quinlan1986induction} (IG) metrics, further support interpretable and trustworthy decision-making by highlighting the contributions of different features and the confidence in decisions. In this study, we integrate these techniques into our NSAI framework to assess and enhance predictive reliability in uncertain conditions.

\section{Methodology}
\label{sec:methodology}
The architecture of the ODXU model is shown in Figure~\ref{fig:pipeline}. The raw network payload is first processed using the Payload-Byte tool, which extracts the 1500 bytes of each packet to form fixed-length feature vectors. These 1500-byte feature vectors are then passed into a DEC model, which encodes them into 12-dimensional latent representations.

The latent representations produced by the DEC serve as input to two downstream models: (1) an XGBoost classifier for attack recognition and (2) a metamodel for UQ, which evaluates the uncertainty of the XGBoost predictions.

The DEC model is trained in two phases, as shown in Figure~\ref{fig:dec_model}: \textit{I) AutoEncoder (AE) pretraining:} In this phase, an AE is trained to map the input data in lower-dimensional features by using an encoder and reconstruct the input data by using a decoder \cite{xie2016dec}.
\textit{II) Clustering training:} In this phase, the decoder is replaced by a clustering model, and the parameters of the clustering model are optimized while the encoder remains fixed. This phase adjusts the cluster centroids and the soft assignments of data points to clusters, refining the latent space to improve the separability of the clusters \cite{xie2016dec}.

\subsection{Transfer Learning Framework for ODXU}
To assess the transferability of the ODXU model across different datasets or tasks, we asked two research questions:
\begin{enumerate}
    \item Which components of the ODXU architecture (e.g., AE, clustering, XGBoost) should be fine-tuned or trained for effective transfer learning?
    \item How many labeled samples are required to outperform baseline machine learning models such as Fully Connected Neural Networks (FcNN) or 1D Convolutional Neural Networks (1D-CNN)?
\end{enumerate}

\begin{table}[ht!]
    \caption{Transfer Learning Scenarios; FT: fine-tune}
    \label{tab:scenarios}
    \centering
    \renewcommand{\arraystretch}{1.15}
    \begin{tabularx}{0.75\linewidth}{>{\hsize=1\hsize}X*{6}{>{\centering\arraybackslash}p{1cm}}}
    \hline
    \multirow{2}{*}{\textbf{ODXU Components}} & \multicolumn{6}{c}{\textbf{Case}} \\
    \cline{2-7}
    & \textbf{1} & \textbf{2} & \textbf{3} & \textbf{4} & \textbf{5} & \textbf{6} \\
    \hline
    AE                   & FT     & As is  & As is  & FT     & As is  & As is  \\
    Clustering           & Train  & FT     & Train  & Train  & FT     & Train  \\
    Classifier (XGBoost) & Train  & Train  & Train  & FT     & FT     & FT     \\
    \hline
    \end{tabularx}
\end{table}

To answer these questions, we designed an experiment using the six transfer learning scenarios as shown in Table~\ref{tab:scenarios}.
The AE has two options: ``As is,'' where the pre-trained AE from CIC-IDS-2017 is loaded and used without any further training, and ``FT'' (fine-tune), where the pre-trained AE is loaded and then further trained on a target dataset. The clustering component also considers two options. In the ``FT'' option, it loads a pre-trained clustering model from CIC-IDS-2017 and fine-tunes it on the target dataset. In the ``Train'' option, it loads the parameters of the pre-trained AE instead of the pre-trained clustering model and trains these parameters on the target dataset\footnote{\textcolor{black}{Note: As depicted in Figure~\ref{fig:dec_model}, the clustering module is initialized either from its pre-trained parameters of CIC-IDS-2017 or from the encoder of the AE. Therefore, if the AE is fine-tuned, the clustering cannot be initialized from its pre-trained checkpoint, as the encoder has changed. As a result, combinations such as FT-FT-Train or FT-FT-FT are invalid and excluded from our experiments.}}.
The classifier has two options: ``FT,'' where a pre-trained classifier is loaded and fine-tuned on the target dataset, and ``Train,'' where the classifier is trained entirely from scratch.

We adopt six metrics, including multiclass classification accuracy, binary classification accuracy, misclassified positive rate, false omission rate, F1 score, and competence. These metrics are consistent with some prior work \cite{sander2024uncertainty}, which comprehensively evaluates the effectiveness of transfer learning in each case.

\subsection{Uncertainty Quantification Methods}

We assess five uncertainty quantification (UQ) methods for our models. This includes two score-based methods: Confidence Scoring and Shannon Entropy \cite{smith2011quantifying}, as well as the metamodel-based methods.

The UQ, through metamodeling, utilizes a base model $b(.)$ and a metamodel binary classifier $m(.)$, trained to estimate the correctness of $b(.)$'s predictions. The metamodel input, $\boldsymbol{X_\text{MetaUQ}}$, includes the original features $\boldsymbol{X_b}$ and augmented features derived from the base model. The training target $\boldsymbol{y_m}$, which denotes whether the base model predicts the sample's labels correctly (0) or not (1), is defined as:
\begin{align}
\label{target}
\boldsymbol{y_m} = 
\begin{cases} 
0 & \text{if }  b(x_b) = y_b\\
1 & \text{if } b(x_b) \neq y_b ,
\end{cases}
\end{align}
where $y_b$ is the true label of $x_b$ and $b(x_b)$ is the predicted output from the base model.

The metamodel, $m(.)$, will be trained using the target variable in Eq. \eqref{target} to minimize the loss between the training target $\boldsymbol{y_m}$ and the output $m(x_m) = z$. As a result, the output of the metamodel, $z$, provides an estimate of the probability that the base model’s prediction $b(x_b) = y_b$ is correct, giving us a measure of the model’s confidence in its predictions. A general interface of the metamodel is shown in Figure \ref{fig:metamodel}.

\subsubsection{Confidence scoring} is a straightforward and efficient method for estimating a model's certainty. The certainty scores can be computed using the order statistic of the prediction probabilities following the formulation:
\begin{equation}
    z_\text{conf}(x) = p_{(k)} - p_{(k-1)},
    \label{eqn:confidence}
\end{equation}
where $z_\text{conf}(x)$ is the certainty score of a sample $x$, $ p_{(k)}$ is the largest probability from the list of probabilities $\boldsymbol{p}$ of $k$ possible outcomes. A higher score indicates greater certainty, while a smaller value suggests more uncertainty. 

\subsubsection{Shannon entropy} is a well-known concept from information theory and is commonly used to measure uncertainty. This approach calculates the entropy for each sample based on the predicted probabilities for all classes generated by the base model. The entropy for a given sample $x$ is computed using the following formula:
\begin{equation}
    z_\text{entropy}(x) = -\sum{p_{i} \times log(p_{i})},
    \label{eqn:entropy}
\end{equation}
where $p_i$ represents the probability of class $i$, and the sum is taken over all possible classes. Unlike the confidence score, a higher entropy value indicates more uncertainty in the model’s prediction, while a lower entropy value suggests higher confidence.
 
\subsubsection{MetaUQ} is a metamodel approach to UQ. Three metamodels were considered, each utilizing a distinct type of uncertainty scores to augment the base model data. 

\paragraph{$\text{MetaUQ}_{prob}$} The metamodel is augmented with the sorted predicted probabilities, $\boldsymbol{p'}$, and the confidence score (Eq. \eqref{eqn:confidence}). Including $\boldsymbol{p'}$ provides a more comprehensive view of the model's predictive distribution \cite{tagasovska2019single}. While the confidence score captures only the two top class probabilities, the full sorted vector reflects the distribution of belief across all classes. This enables the metamodel to identify uncertainty patterns, such as class ambiguity or flat distributions, that are not captured by the input $\boldsymbol{X_b}$ or confidence score alone. The augmented input of the metamodel ($\text{MetaUQ}_{prob}$)  can be written as:
\begin{equation}
    \boldsymbol{X}_{\text{MetaUQ}_{prob}} = [\boldsymbol{X_b}, \boldsymbol{p}', \boldsymbol{z_\text{conf}}].
\end{equation}

\paragraph{$\text{MetaUQ}_{SHAP}$} Originally, Shapley Additive Explanations (SHAP) \cite{lundberg2017unified} values were a game-theoretic measure to attribute value to members of a coalition. Still, recently they have been widely adopted in machine learning for explainability purposes \cite{lundberg_shap}. In this context, we consider the individual features in $x$ as members of a coalition and aim to attribute to each feature a value that represents its contribution to the final prediction made by the model.

For tree-based models, such as XGBoost, SHAP values can be computed efficiently using a polynomial-time algorithm that leverages the structure of decision trees \cite{lundberg2020local}. This allows us to get local explanations (i.e., the contribution of each feature to an individual prediction) without relying on approximations 
or sampling. In our implementation, for each input instance $\boldsymbol{x}$, we compute SHAP values as follows \cite{lundberg2017unified}:
\begin{equation}
    \phi_i(b, x) = \sum_{S \subseteq I \setminus \{i\}} \frac{|S|! (|I|-|S|-1)!}{|I|!} \left[b(S \cup \{i\}) - b(S)\right],
    \label{eqn:shapvalue}
\end{equation}
where
$\phi_i(b,x)$ is the SHAP value for feature $i$ with respect to the base model $b(.)$ and input instance $\boldsymbol{x}$. $!$ is the factorial operator, $|\cdot|$ denotes the number samples of set or subsets.
$I$ is the set of input features. $S \subseteq I \setminus \{i\}$ is a subset of features from $I$ excluding feature $i$.
$b(S \cup \{i\}) - b(S)$ represents the change in the model's output when feature $i$ is added to subset $S$, compared to the output when only $S$ is used.
We then extract SHAP values for the predicted class only, following the practices in class-specific interpretability \cite{lundberg_shap}. 

The final augmented input to the SHAP metamodel is:
\begin{equation}
    \boldsymbol{X}_{\text{MetaUQ}_{SHAP}} = [\boldsymbol{X_b}, \boldsymbol{\phi}(b,x)].
    \label{eqn:shap_augment}
\end{equation}

\paragraph{$\text{MetaUQ}_{IG}$} Information Gain (IG) is a key concept in decision trees and gradient boosting, and it measures how much a split on a feature reduces uncertainty in the prediction. For classical classification tasks, the IG of a feature is calculated as follows: \cite{quinlan1986induction}:
\begin{equation}
    \text{IG}(I, f) = H(I) - H(I|f),
\end{equation}
where $H(I)$ represents the entropy of the entire dataset $I$, and $H(I|f)$ represents the conditional entropy, measuring the remaining uncertainty in the dataset $I$ after splitting $I$ based on the values of feature $f$.

In XGBoost, the concept of IG is extended to general loss functions. In this context, IG scores are defined as the expected reduction in loss due to a split; more details are here \cite{chen2016xgboost}.

These scores represent the average usefulness of each feature across all decision splits. We then replicate the gain scores across all data samples to form an IG matrix, denoted as $\text{IG}_{\text{matrix}}$. We also add the sorted class probabilities from the base model’s output. As a result, the final augmented input to the metamodel becomes:
\begin{equation}
    \boldsymbol{X}_{\text{MetaUQ}_{IG}} = [\boldsymbol{X_b},\boldsymbol{p'}, \text{IG}_{\text{matrix}}]
    \label{eqn:IG_augment}
\end{equation}

\subsubsection{UQ Evaluation Metrics}

To evaluate the effectiveness of our UQ methods on the transfer learning model, we focus on two tasks: misclassification detection and Open Set Recognition (OSR) detection. 
We compute the Area Under the Receiver Operating Characteristic Curve (AUROC) for each task to detect the misclassified and OSR samples. These two AUROC scores are the core features of evaluating the UQ methods. 

\section{Experimental Setup}
\label{sec:exp_setup}
All experiments in this study were conducted using the Army Cyber Institute (ACI) Internet of Things (IoT) Network Traffic Dataset 2023 (ACI-IoT-2023) \cite{aciiot2023}, which provides recent and comprehensive intrusion scenarios in Internet of Things (IoT) environments. The descriptive statistics of the dataset are shown in Table~\ref{tab:aci_iot2023_class_stats}.

\textcolor{black}{
The experiments were performed on a system with dual Intel® Xeon® Gold 5218R CPUs (2.10 GHz), providing a total of 80 logical threads and 754 GiB of system memory. For GPU acceleration, each experiment was conducted on a single NVIDIA A40 GPU, one of eight available, which is based on the Ampere architecture and features 48 GB of dedicated memory. Additionally, all experiments were timed and monitored to evaluate their computational efficiency.
}

\begin{table}[ht!]
\centering
\caption{Descriptive statistics of class distribution in the ACI-IoT-2023 dataset.}
\label{tab:aci_iot2023_class_stats}
\renewcommand{\arraystretch}{1.15}
\begin{tabular}{|l|r|r|}
\hline
\textbf{Class} & \textbf{Samples} & \textbf{Percent (\%)} \\
\hline
Benign              & 601,868 & 95.31 \\
DNS Flood           & 18,577  & 2.94 \\
Dictionary Attack   & 4,645   & 0.74 \\
Slowloris           & 2,974   & 0.47 \\
SYN Flood           & 2,113   & 0.33 \\
Port Scan           & 582       & 0.09 \\
Vulnerability Scan  & 445       & 0.07 \\
OS Scan             & 156       & 0.02 \\
UDP Flood           & 68        & 0.01 \\
ICMP Flood          & 58        & 0.01 \\
\hline
\textbf{Total}      & \textbf{631,486} & \textbf{100.00} \\
\hline
\end{tabular}
\end{table}

\subsection{Attack Recognition}
To address the class imbalance in the ACI-IoT 2023 dataset, the benign class was downsampled by $95\%$, while the ICMP Flood and UDP Flood attack classes were upsampled by $200\%$ \cite{sander2024uncertainty}. The adjusted dataset was split into two subsets: \emph{DEC-Train} and \emph{DEC-Test}, with a 50/50 split. The \textit{DEC-Train} set was further subsampled into portions of $10\%$, $25\%$, $50\%$, and $75\%$, which were then used to train and evaluate the models. Each portion of \textit{DEC-Train} was divided into training and validation sets (75/25 split). Meanwhile, \textit{DEC-Test} was used for XGBoost training, which is output from DEC and consists of 12 features. The \textit{DEC-Test} set was further split 50/50 into \textit{XGBoost-Train} and \textit{XGBoost-Test}.   

For the evaluation of transfer learning models, we trained two neural-based models: a FcNN and a 1D-CNN. The FcNN model consisted of three hidden layers with sizes [1024, 512, 100], totaling 2,115,180 trainable parameters. The 1D-CNN model consisted of three convolutional layers with channels [32, 64, 128] and a kernel size of [3], followed by a hidden layer with 50 neurons. The 1D-CNN model had a total of 181,904 trainable parameters.

\subsection{Misclassification and Open Set Recognition Detection}
For the misclassification and OSR detection tasks, the Slowloris attack was held out as the ``unknown'' attack. The remaining samples were subsampled to balance the benign and attack classes, with the dataset split following the same procedure as in the attack recognition task.
In the case of our UQ metamodel, a more refined training set was required. Using a simple split of \textit{XGBoost-Test} resulted in poor performance. We found that a metamodel trained on a highly accurate base model would tend to classify all samples as high certainty, which minimized the model's training objective. Through experimentation, we determined that holding five times as Many correctly classified samples as misclassified samples resulted in improved MetaUQ performance. Consequently, we created new labels for the \textit{XGBoost-Test} set: 0 for correct predictions and 1 for incorrect predictions. Class 0 was then subsampled to five times the size of class 1. This adjusted dataset was split 80/20 into \textit{Metamodel-Train} and \textit{Metamodel-Test}. 
For testing, we generated a new test set for the OSR detection task by combining equal numbers of unknown samples and \textit{Metamodel-Test} samples. This combined set was used for the final evaluation of our models.

\section{Results and Discussion}
\label{sec:results}

\subsection{Attacks Recognition}
Table~\ref{tab:vary_dataset} presents the multiclass accuracy of transfer learning models across different training set portions, based on the configurations in Table~\ref{tab:scenarios}. When considering the composition of the DEC components, performance increases from models using a pre-trained AE (As is) with fine-tuned clustering (e.g., Case~2), to those fine-tuning the AE and training the clustering module (e.g., Case~1), and is highest when using the pre-trained AE and training the clustering (e.g., Case~3). For example, with 50\% of the training set, Case~2 achieves .9799 accuracy, Case~1 improves to .9805, and Case~3 reaches the highest at .9827.

\begin{table}[ht!]
\centering
\caption{The multiclass accuracy of transfer learning across varying portions of the training dataset.}
\label{tab:vary_dataset}
\renewcommand{\arraystretch}{1.15}
\begin{tabular}{|c|c|c|c|c|c|c|c|}
\hline
\textbf{Percentage} & \textbf{Case 1} & \textbf{Case 2} & \textbf{Case 3} & \textbf{Case 4} & 
\textbf{Case 5} & 
\textbf{Case 6} \\
\hline
10  & .9666 & .9616 & .9764 & .9680 & .9659 & .9784 \\
\hline
25  & .9690 & .9679 & .9789 & .9736 & .9706 & .9802 \\
\hline
50  & .9805 & .9799 & \textbf{.9827} & \textbf{.9813} & \textbf{.9808} & \textbf{.9841} \\
\hline
75  & \textbf{.9824} & \textbf{.9809} & \textbf{.9836} & \textbf{.9833} & \textbf{.9816} & \textbf{.9845} \\
\hline
\end{tabular}
\end{table}

Furthermore, when comparing models with the same AE and clustering settings, fine-tuning the classifier (Cases~4 to 6) consistently yields higher performance than training it from scratch (Cases~1 to 3). For instance, at 75\% of the data, Case~3 achieves .9836, while Case~6 achieves a higher accuracy of .9845. These results suggest that starting from a well-initialized classifier leads to better generalization and performance, emphasizing the advantage of transfer learning.
When compared with neural-based models (.9808 for FcNN and .9679 for 1D-CNN), the transfer learning models can outperform them when trained with at least 50\% ($\approx$16,000 samples) of the dataset in Cases 3--6, and at least 75\% ($\approx$ 23,000 samples) in Cases 1 and 2\footnote{\textcolor{black}{These results are based on an ablation study and examining several cases.}}. \textcolor{black}{As Case~6 (AE: As is, clustering: Train, and classifier: FT) provides the highest accuracy, we will use Case~6 for further experimentation in subsequent experiments.}

\paragraph{Early Stopping Improvement}
During the experiments, all models were trained in the same setting (i.e., number of epochs). The models improve quickly at first, then slow down. To prevent slow progress, we introduced ``early stopping rounds'' ($\eta$) and two thresholds in loss ($\delta$) as hyperparameters. We assume that if the change in the loss of AE pre-training or clustering training within the $\delta$ for a specified number of epochs, the model stops. We tested the Case~6 model using early stopping rounds of [10, 15, 20], and $\delta$ values for AE pre-training of [0.0005, 0.001] and clustering training of [0.005, 0.01], with 50\% and 75\% dataset portions.

\begin{table}[ht!]
\centering
\caption{The multiclass accuracy across hyperparameters.}
\label{tab:finetune_config}
\renewcommand{\arraystretch}{1.15}
\begin{tabularx}{0.95\linewidth}{|c|c|*{2}{Y|}*{2}{Y|}*{2}{Y|}}
\hline
\multirow{2}{*}{\textbf{Exp}} & \multirow{2}{*}{$\bm{\eta}$} &
\multicolumn{2}{c|}{$\bm{\delta}$} &
\multicolumn{2}{c|}{\textbf{Accuracy}} &
\multicolumn{2}{c|}{\textbf{Training time}} \\
\cline{3-8}
 &  & \textbf{AE} & \textbf{Cluster} & \textbf{50\%} & \textbf{75\%} & \textbf{50\%} & \textbf{75\%} \\
\hline
1 & 10 & 0.001  & 0.01  & .9807 & .9791 & 0:23:07 & 0:20:09 \\
\hline
2 & 10 & 0.0005 & 0.005 & \textbf{.9817} & \textbf{.9820} & 0:25:00 & 0:28:48 \\
\hline
3 & 15 & 0.001  & 0.01  & \textbf{.9815} & .9806 & 0:27:32 & 0:21:38 \\
\hline
4 & 15 & 0.0005 & 0.005 & \textbf{.9819} & \textbf{.9831} & 0:29:10 & 0:37:01 \\
\hline
5 & 20 & 0.001  & 0.01  & \textbf{.9820} & \textbf{.9829} & 0:39:02 & 0:42:46 \\
\hline
6 & 20 & 0.0005 & 0.005 & \textbf{.9824} & \textbf{.9834} & \textbf{0:49:39} & \textbf{1:02:50} \\
\hline
\end{tabularx}
\end{table}

Table~\ref{tab:finetune_config} presents the multiclass accuracy across various combinations of hyperparameters. The ``\textbf{Exp}'' column denotes the experiment index. For the 50\% training data, all configurations except Experiment~1 outperform the FcNN baseline (.9808). The best accuracy for 50\% data is obtained in Experiment~6 with an accuracy of .9824. For the 75\% data portion, Experiments~2, 4, 5, and 6 outperform the FcNN baseline, with Experiment~6 again achieving the highest accuracy of .9834. These results confirm that Experiment~6 offers the most robust configuration overall.

In addition to performance, training time is an important consideration. As $\eta$ increases and $\delta$ decreases, training time, \textcolor{black}{showing in hh:mm:ss format}, tends to rise. For example, Experiment~6, which combines a higher $\eta$ (20) and lower $\delta$ values (0.0005, 0.005), requires the longest training time for both data portions (up to \textbf{1:02:50} for 75\%).
The choice of specific hyperparameters and portions of the dataset will depend on specific tasks or the dataset.
In our study, the next experiment will proceed with Case~6 using the hyperparameters from Experiment~6. The extended evaluation of this configuration on the attack recognition task is reported in Table~\ref{tab:ACI_compare}.

\begin{table}[ht!]
\centering
\caption{Results of six metrics across models.}
\label{tab:ACI_compare}
\renewcommand{\arraystretch}{1.15}
\begin{tabular}{|c|c|c|c|}
\hline
\textbf{Measurement} & \textbf{FcNN} & \textbf{1D-CNN} & \textbf{Case 6} \\
\hline
Multiclass Accuracy        & .981 & .968 & \textbf{.983} \\
\hline
Binary Accuracy            & .985 & .974 & \textbf{.987} \\
\hline
Misclassified Positive Rate & .022 & .035 & \textbf{.019} \\
\hline
False Omission Rate        & .016 & .029 & \textbf{.014} \\
\hline
F1 Score                   & .985 & .974 & \textbf{.988} \\
\hline
Competence                 & .948 & .935 & \textbf{.969} \\
\hline
\end{tabular}
\end{table} 

\subsection{Misclassification and Open Set Detection}

The results for misclassification and OSR detection are shown in Table \ref{table:obj2_aci}. In misclassification detection, Confidence and Shannon Entropy perform similarly, both achieving an accuracy of .911, while the metamodels obtain higher accuracy. The $\text{MetaUQ}_{prob}$ and $\text{MetaUQ}_{SHAP}$ have accuracy values close to .924, and the $\text{MetaUQ}_{IG}$ achieved the highest performance at .926. A similar trend is observed with TP@(TN=.95). These results suggest that the $\text{MetaUQ}_{IG}$ is effective for detecting misclassification.

\begin{table}[H]
\caption{Performance comparison across UQ methods.}
\centering
\label{table:obj2_aci}
\renewcommand{\arraystretch}{1.2}
\begin{tabular}{|c|c|c|c|c|c|}
\hline
\multirow{2}{*}{\textbf{Metrics }} & \multicolumn{2}{c|}{\textbf{Score-based}}  & \multicolumn{3}{c|}{\textbf{MetaUQ}}   \\
\cline{2-6}
\textbf{} & \textbf{Confidence} & \textbf{Entropy} & \textbf{Prob} & \textbf{SHAP} & \textbf{IG} \\
\hline
Misclassification AUROC & .911  & .911  & .924  & .924  & \textbf{.926}  \\
\hline
Misclassification TP@(TN=.95) & .529  & .529 & .559& .559  & \textbf{.588} \\
\hline
Unknown attack AUROC & .916 & .919 & .921 & \textbf{.938} & .921 \\
\hline
Unknown attack TP@(TN=.95) & .435 & .462 & .489 & \textbf{.590} & .469 \\
\hline
\end{tabular}
\end{table}

For OSR detection, there is little difference between Confidence, Shannon Entropy, $\text{MetaUQ}_{prob}$, and $\text{MetaUQ}_{IG}$. However, the $\text{MetaUQ}_{SHAP}$ shows a big gap in AUROC for unknown attack detection. When fixing the true negative rate at 95\%, the $\text{MetaUQ}_{SHAP}$ exceeds the second-highest method ($\text{MetaUQ}_{prob}$) by more than 10\%. This indicates that the $\text{MetaUQ}_{SHAP}$ metamodel provides the best performance for OSR detection.

\section{Conclusion}
\label{sec:conclusion}
This paper presents a transfer learning framework for the ODXU Neurosymbolic AI model applied in network intrusion detection systems. 
The results indicate that utilizing a pre-trained AE, followed by retraining the clustering algorithm and fine-tuning the classifier model (XGBoost), yields the highest accuracy. Transfer learning models began to surpass neural-based models when trained on at least 50\% of the data, or 16,000 samples (these findings are based on the ablation study when examining several cases). To prevent prolonged training times, we implemented an early stopping condition that halts training if the AE and clustering losses drop below 0.0005 and 0.005, respectively, for 20 consecutive epochs.

Our findings also suggest that metamodel-based methods are more effective than score-based methods for uncertainty quantification (UQ), with each metamodel being tailored for specific tasks. However, we recognize that these results may differ depending on the datasets or tasks used. Therefore, future work will focus on applying our transfer learning model to additional cybersecurity datasets, including the Canadian Institute for Cybersecurity (CIC) IoT 2023 dataset and the Unified Multimodal Network Intrusion Detection Systems (UM-NIDS) dataset, to further assess its performance.

\section*{Acknowledgment}

This work was supported by the U.S. Military Academy (USMA) under Cooperative Agreement No. W911NF-23-2-0108 and the Defense Advanced Research Projects Agency (DARPA) under Support Agreement No. USMA 23004. The views and conclusions expressed in this paper are those of the authors and do not reflect the official policy or position of the U.S. Military Academy, U.S. Army, U.S. Department of Defense, or U.S. Government.

\bibliographystyle{unsrt}   
\bibliography{main}

\begin{thebibliography}{10}

\bibitem{al2021intelligent}
Mohammad Al-Omari, Majdi Rawashdeh, Fadi Qutaishat, Mohammad Alshira’H, and Nedal Ababneh.
\newblock An intelligent tree-based intrusion detection model for cyber security.
\newblock {\em Journal of Network and Systems Management}, 29(2):20, 2021.

\bibitem{sander2024uncertainty}
Jacob Sander, Chung-En~Johnny Yu, Brian Jalaian, and Nathaniel~D. Bastian.
\newblock Uncertainty-quantified neurosymbolic ai for open set recognition in network intrusion detection.
\newblock In {\em 2024 IEEE Military Communications Conference (MILCOM)}, pages 13--18, Washington, DC, USA, 2024. IEEE.

\bibitem{jalaian2023neurosymbolic}
Brian Jalaian and Nathaniel~D. Bastian.
\newblock Neurosymbolic ai in cybersecurity: Bridging pattern recognition and symbolic reasoning.
\newblock In {\em Proceedings of the 2023 IEEE Military Communications Conference (MILCOM)}, pages 268--273, Boston, MA, USA, 2023. IEEE.

\bibitem{shahoveisi2023application}
Fereshteh Shahoveisi, Hamed Taheri~Gorji, Seyedmojtaba Shahabi, Seyedali Hosseinirad, Samuel Markell, and Fartash Vasefi.
\newblock Application of image processing and transfer learning for the detection of rust disease.
\newblock {\em Scientific Reports}, 13(1):5133, 2023.

\bibitem{amiriparian2022deepspectrumlite}
Shahin Amiriparian, Tobias H{\"u}bner, Vincent Karas, Maurice Gerczuk, Sandra Ottl, and Bj{\"o}rn~W Schuller.
\newblock Deepspectrumlite: A power-efficient transfer learning framework for embedded speech and audio processing from decentralized data.
\newblock {\em Frontiers in Artificial Intelligence}, 5:856232, 2022.

\bibitem{moon2014multimodal}
Seungwhan Moon, Suyoun Kim, and Haohan Wang.
\newblock Multimodal transfer deep learning with applications in audio-visual recognition.
\newblock \url{https://arxiv.org/abs/1412.3121}, 2014.
\newblock arXiv:1412.3121 [cs.LG].

\bibitem{kim2022transfer}
Hee~E Kim, Alejandro Cosa-Linan, Nandhini Santhanam, Mahboubeh Jannesari, Mate~E Maros, and Thomas Ganslandt.
\newblock Transfer learning for medical image classification: a literature review.
\newblock {\em BMC medical imaging}, 22(1):69, 2022.

\bibitem{sharafaldincicids2017}
Iman Sharafaldin, Arash~Habibi Lashkari, Ali~A Ghorbani, et~al.
\newblock Toward generating a new intrusion detection dataset and intrusion traffic characterization.
\newblock {\em ICISSp}, 1:108--116, 2018.

\bibitem{lee2022surgical}
Yoonho Lee, Annie~S. Chen, Fahim Tajwar, Ananya Kumar, Huaxiu Yao, Percy Liang, and Chelsea Finn.
\newblock Surgical fine-tuning improves adaptation to distribution shifts.
\newblock \url{https://arxiv.org/abs/2210.11466}, 2022.
\newblock arXiv:2210.11466 [cs.LG].

\bibitem{liu2021transtailor}
Bingyan Liu, Yifeng Cai, Yao Guo, and Xiangqun Chen.
\newblock Transtailor: Pruning the pre-trained model for improved transfer learning.
\newblock In {\em Proceedings of the AAAI Conference on Artificial Intelligence}, volume~35, pages 8627--8634, Vancouver, BC, Canada (held virtually), 2021. AAAI Press.

\bibitem{guo2019spottune}
Yunhui Guo, Honghui Shi, Abhishek Kumar, Kristen Grauman, Tajana Rosing, and Rogerio Feris.
\newblock Spottune: Transfer learning through adaptive fine-tuning.
\newblock In {\em Proceedings of the IEEE/CVF Conference on Computer Vision and Pattern Recognition (CVPR)}, pages 4805--4814, Long Beach, CA, USA, 2019. IEEE.

\bibitem{smith2011quantifying}
Geoffrey Smith.
\newblock Quantifying information flow using min-entropy.
\newblock In {\em Proceedings of the 2011 Eighth International Conference on Quantitative Evaluation of Systems (QEST)}, pages 159--167, Aachen, Germany, 2011. IEEE.

\bibitem{chen2019ibmwhitebox}
Tongfei Chen, Jiří Navrátil, Vijay Iyengar, and Karthikeyan Shanmugam.
\newblock Confidence scoring using whitebox meta-models with linear classifier probes.
\newblock In {\em Proceedings of the 22nd International Conference on Artificial Intelligence and Statistics (AISTATS 2019)}, pages 1467--1475, Naha, Okinawa, Japan, 2019. Proceedings of Machine Learning Research.

\bibitem{lundberg2017unified}
Scott~M. Lundberg and Su-In Lee.
\newblock A unified approach to interpreting model predictions.
\newblock {\em Advances in Neural Information Processing Systems}, 30:4765--4774, 2017.

\bibitem{quinlan1986induction}
J.~Ross Quinlan.
\newblock Induction of decision trees.
\newblock {\em Machine learning}, 1:81--106, 1986.

\bibitem{xie2016dec}
Junyuan Xie, Ross Girshick, and Ali Farhadi.
\newblock Unsupervised deep embedding for clustering analysis.
\newblock In {\em Proceedings of the 33rd International Conference on Machine Learning (ICML)}, ICML'16, pages 478--487, New York, NY, USA, 2016. JMLR.org.

\bibitem{tagasovska2019single}
Natasa Tagasovska and David Lopez-Paz.
\newblock Single-model uncertainties for deep learning.
\newblock {\em Advances in Neural Information Processing Systems}, 32:6415--6425, 2019.

\bibitem{lundberg_shap}
Scott~M. Lundberg, Gabriel~G. Erion, and Su-In Lee.
\newblock Consistent individualized feature attribution for tree ensembles, 2019.

\bibitem{lundberg2020local}
Scott~M Lundberg, Gabriel Erion, Hugh Chen, Alex DeGrave, Jordan~M Prutkin, Bala Nair, Ronit Katz, Jonathan Himmelfarb, Nisha Bansal, and Su-In Lee.
\newblock From local explanations to global understanding with explainable ai for trees.
\newblock {\em Nature machine intelligence}, 2(1):56--67, 2020.

\bibitem{chen2016xgboost}
Tianqi Chen and Carlos Guestrin.
\newblock Xgboost: A scalable tree boosting system.
\newblock In {\em Proceedings of the 22nd ACM SIGKDD International Conference on Knowledge Discovery and Data Mining (KDD '16)}, pages 785--794, San Francisco, CA, USA, 2016. ACM.

\bibitem{aciiot2023}
Nathaniel Bastian, David Bierbrauer, Morgan McKenzie, and Emily Nack.
\newblock {ACI IoT Network Traffic Dataset 2023}.
\newblock \url{https://dx.doi.org/10.21227/qacj-3x32}, 2023.
\newblock DOI: 10.21227/qacj-3x32.

\end{thebibliography}

\end{document}